%% file: sample-sigconf.tex
\documentclass[sigconf]{acmart}

\usepackage{graphicx}
\usepackage{tikz}
\usepackage{booktabs}
\usepackage{hyperref}
\usepackage{url}
\usepackage{xspace}
\usepackage{amsmath}
\usepackage{tabularx}
\DeclareRobustCommand\circled[1]{\tikz[baseline=(char.base)]{
            \node[shape=circle,draw,inner sep=1pt] (char) {#1};}}
\newcommand{\mask}{\texttt{\small [MASK]\xspace}}
\newcommand{\pad}{\texttt{\small [PAD]\xspace}}
\newcommand{\sos}{\texttt{\small [SOS]\xspace}}
\newcommand{\eos}{\texttt{\small [EOS]\xspace}}

\setcopyright{acmcopyright}

\acmDOI{xx.xxx/xxx_x}

\acmISBN{978-1-4503-8713-2/22/04}

\acmYear{2022}
\copyrightyear{2022}

\acmArticle{4}
\acmPrice{15.00}

\setcopyright{none}
\renewcommand\footnotetextcopyrightpermission[1]{}
\begin{document}
\title{What Averages Do Not Tell}
\subtitle{Predicting Real Life Processes with Sequential Deep Learning}

\renewcommand{\shorttitle}{Sequential Deep Learning Models}

 \author{Istv\'an Ketyk\'o}
 \orcid{0000-0003-4931-4580}
 \affiliation{%
   \institution{Eindhoven University of Technology, Mathematics and Computer Science}
   \city{Eindhoven}
   \state{the Netherlands}
 }
 \email{i.ketyko@tue.nl}

 \author{Felix Mannhardt}
 \orcid{0000-0003-4931-4580}
 \affiliation{%
   \institution{Eindhoven University of Technology, Mathematics and Computer Science}
   \city{Eindhoven}
   \state{the Netherlands}
 }
 \email{f.mannhardt@tue.nl}

 \author{Marwan Hassani}
 \orcid{0000-0003-4931-4580}
 \affiliation{%
   \institution{Eindhoven University of Technology, Mathematics and Computer Science}
   \city{Eindhoven}
   \state{the Netherlands}
 }
 \email{m.hassani@tue.nl}

 \author{Boudewijn F. van Dongen}
 \orcid{0000-0003-4931-4580}
 \affiliation{%
   \institution{Eindhoven University of Technology, Mathematics and Computer Science}
   \city{Eindhoven}
   \state{the Netherlands}
 }
 \email{b.f.v.dongen@tue.nl}

\renewcommand{\shortauthors}{I. Ketyk\'o et al.}

\begin{abstract}
Deep Learning is proven to be an effective tool for modeling sequential data as shown by the success in Natural Language, Computer Vision and Signal Processing.
Process Mining concerns discovering insights on business processes from their execution data that are logged by supporting information systems.
The logged data (event log) is formed of event sequences (traces) that correspond to executions of a process.
Many Deep Learning techniques have been successfully adapted for predictive Process Mining that aims to predict process outcomes, remaining time, the next event, or even the suffix of running traces.
Traces in Process Mining are multimodal sequences and very differently structured than natural language sentences or images. This may require a different approach to processing.
So far, there has been little focus on these differences and the challenges introduced. Looking at suffix prediction as the most challenging of these tasks, the performance of Deep Learning models was evaluated only on average measures and for a small number of real-life event logs. Comparing the results between papers is difficult due to different pre-processing and evaluation strategies. Challenges that may be relevant are the skewness of trace-length distribution and the skewness of the activity distribution in real-life event logs. We provide an end-to-end framework which enables to compare the performance of seven state-of-the-art sequential architectures in common settings. Results show that sequence modeling still has a lot of room for improvement for majority of the more complex datasets. Further research and insights are required to get consistent performance not just in average measures but additionally over all the prefixes.
\end{abstract}

%
%



\maketitle

\section{Introduction}

Process Mining (PM) is a recent research field that advances Business Process Management (BPM) using Machine Learning (ML) and data mining solutions. Two first class citizens in this field are event logs extracted from (business) information systems, and explainable process models. The event logs contain sequential end-to-end record of what happened in the \emph{cases} which are instances of the process. A process discovery algorithm extracts a process model out of such an event log. A process model is a graphical representation of the process behavior and define which sequences of process steps (activities) are possible. A process discovery algorithm extracts such a process model from the sequential process behavior observed as event sequences (traces) in reality and as recorded in an event log.

Learning to predict how active instances or cases of a business process are likely to unfold in the future is the main goal of predictive process mining or predictive process monitoring~\cite{Neu2021}. Fueled by the advances made in Deep Learning (DL), the prospect of accurate predictors of the (process) future has received a lot of attention and promises organizations to act on process problems before they manifest. This complements the more retrospective nature of process discovery, which provides aggregated and interpretable models of historical cases by leveraging events logged by information systems.

Several prediction tasks have been investigated~\cite{Neu2021}: looking at the immediate future by predicting properties of the next event (\emph{next-event prediction}), looking at what happens at the end of the process execution by predicting properties of the case (\emph{outcome prediction}, e.g., loan application acceptance, or customer churn prediction), or attempting to construct the whole process execution by predicting properties of all future events (\emph{suffix prediction}, i.e. sequence generation). Event properties of interest include the event label or activity label and the timestamp. Case or outcome properties are often concerned with some performance indicator of the process. We focus on suffix prediction as the most challenging task for predicting event properties.

DL has been proposed for all three categories of prediction tasks including suffix prediction and consistently outperformed all alternatives based on the remaining methods~\cite{Neu2021}. Many of the DL architectures have been adapted for event logs from Natural Language Processing (NLP) and Computer Vision, and led to substantial improvements. However, traces are multi modal, e.g., contains both label and time, do not form a continuous corpus as in NLP, and the trace-length distribution of the event log can be highly skewed.
DL provides efficient training due to the graphic cards and parallel processing. To ideally enable parallel processing, batching of equal-length samples is applied, which usually needs padding. Trace-length skewness leads to the necessity of introducing a large number of padding tokens.

Two studies have explored how some of the peculiarities of event logs are related to the prediction performance for next-event prediction~\cite{DBLP:journals/dss/HeinrichZJB21} and for outcome prediction~\cite{DBLP:journals/bise/KratschMRS21}. The performance of suffix prediction was evaluated only on average performance measures for a few real-life event logs~\cite{Neu2021}. In addition, different pre-processing and evaluation strategies have been used which makes it difficult to compare the results of the different approaches~\cite{Neu2021,DBLP:journals/corr/abs-2107-01905}.
Our work contributes a unified framework for case suffix prediction with seven sequential DL models, multimodal event attribute fusion, and multi-target prediction (Section~\ref{sec:framework}). The end-to-end framework automatically processes event logs and avoids feature engineering. We show in detail how the performance behavior of different models vary over all prefixes of different length and highlight corresponding challenges (Section~\ref{sec:results}). We show that skewness affects all DL models and that it is insufficient to solely rely on average performance measures.

\section{Problem Formulation and Notation}
In the suffix and remaining time prediction, the dataset, i.e., event log, is a set of sequences (process executions or \emph{traces}) $D = \{\sigma^{(1)},\allowbreak \dots,\sigma^{(d)} \}$, where $d$ is the size of dataset. The $i$-th process execution $\sigma^{(i)} = \langle e_1,\dots,e_n \rangle$ contains a sequence of $n$ events, i.e., $|\sigma^{(i)}| =n$. An event $e_j = (a_j,t_j)$ has two attributes where the former is the event's label, i.e., an activity, and the latter is the event's duration time, or the required execution time. For a given process execution $\sigma^{(i)}$=$\langle (a_1,t_1),\dots, (a_n,t_n) \rangle$, the prefix of events of length $k$ is defined by $\sigma^{(i)}_{\leq k}$=$\langle (a_1,t_1),\dots,(a_k,t_k) \rangle$ and its corresponding suffix of events is $\sigma^{(i)}_{> k}$=$\langle (a_{k+1}, t_{k+1}),\dots, (a_{n}, t_{n}) \rangle$.

\begin{definition}[Suffix prediction and remaining time prediction]
Suppose that there are pairs sample of sequences

\noindent$\mathcal{S}=\allowbreak\{(\sigma_{\leq k}^{(i)}, \sigma_{>k}^{(i)})\}_{i=1}^{i=n}$, where $2 \leq k<|\sigma^{(i)}|$ is the prefix length and $n$ is the sample size (also the set of prefixes $\mathcal{P} = \{\sigma_{\leq k}^{(i)}\}_{i=1}^{i=n}$).
Given a prefix of events sequence, the output prediction is the sequence of events $\hat{\sigma}_{>k} = \langle (a_{k+1},t_{k+1}),\dots,\eos{} \rangle$, where \eos{} is a special symbol added to the end of each process execution to mark the end of the sequence in prepossessing time.
Suffix prediction is the sequence of activities in $\hat{\sigma}_{>k}$, i.e., $\langle a_{k+1},\dots,\eos{} \rangle$. The remaining time prediction is the sum of the predicted duration time $t$ in $\hat{\sigma}_{>k}$, i.e., $\sum t_i$, where $t_i \in \hat{\sigma}_{>k}$.
\label{def:suffix_pred_definition}
\end{definition}

\section{Proposed Framework}
\label{sec:framework}
\begin{figure}[t]
	\centerline{\includegraphics[width=\columnwidth,keepaspectratio]{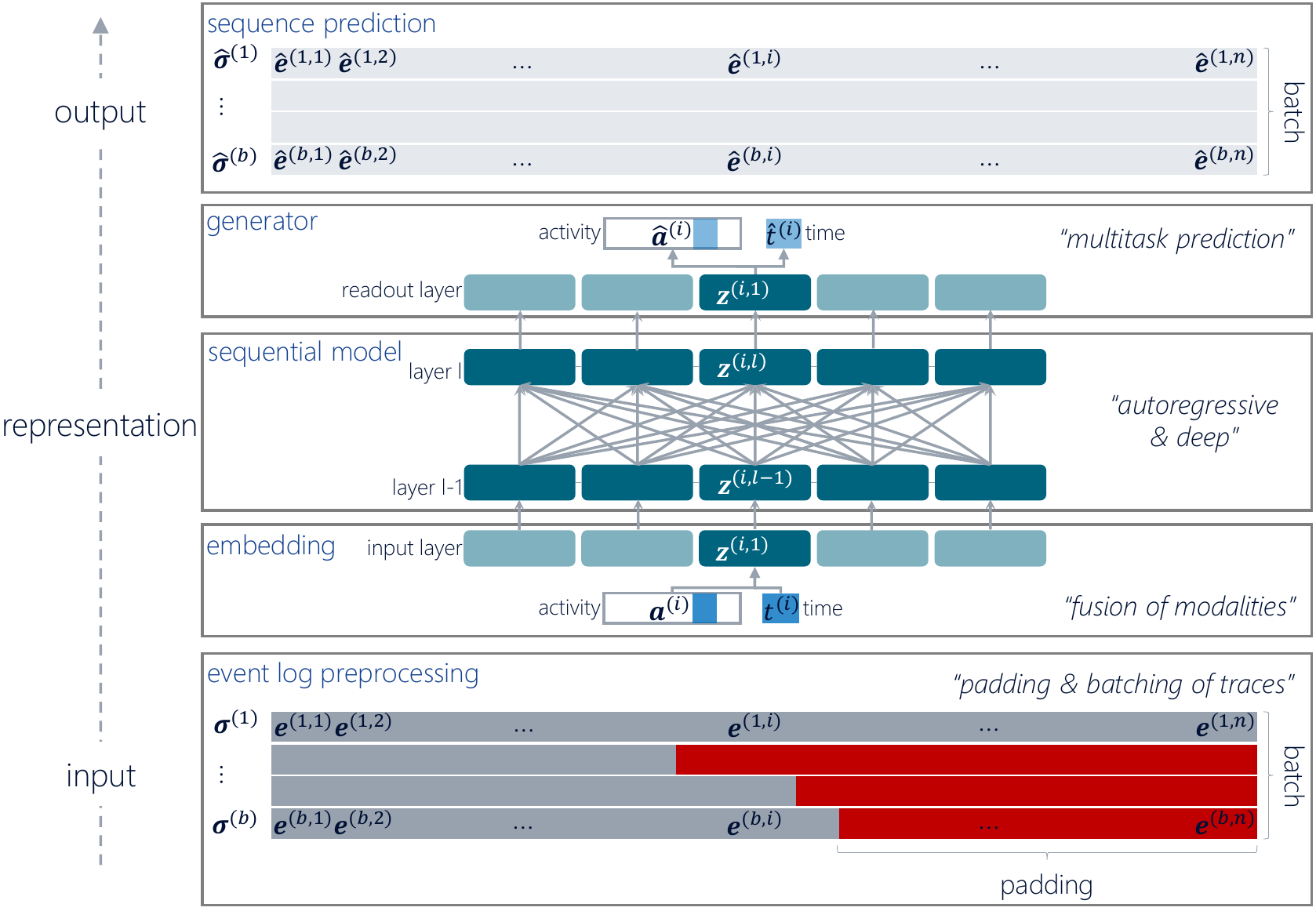}}
	\caption{Unified framework for sequential DL model comparisons.}
	\label{figure:framework}
\end{figure}
The proposed framework in Fig.~\ref{figure:framework} provides deep (i.e. several layers stacked), autoregressive modeling for multi-objective prediction tasks of multimodal sequences (e.g. log traces). Our vision is to perform no feature engineering and to use the original event logs without excluding traces or trace parts, e.g., we do not trim very long nor exclude very short traces.

We describe the three main components of our framework: embedding of an input sequence for the fusion of modalities (Section~\ref{section:embedding}), the pluggable sequential model component (Section~\ref{section:sequential}), and the generator providing multitask predictions (Section~\ref{section:generator}).

\subsection{Embedding} \label{section:embedding}
In the proposed framework each event $e_i = (a_i, t_i)$ is shown by tuple $\mathbf{e}^{(i)} = (\mathbf{a}^{(i)}, t_i)$, where $\mathbf{a}^{(i)}$ is the one-hot encoding of activity $a_i$. We denote the $j$-th entry of a vector by the corresponding subscript, e.g., $\mathbf{a}^{(i)}_j$.
The embedding component maps the input features $\mathbf{e}^{(i)}$ into the latent space of the model.

First, there are two tasks for preprocessing the input event traces.
\noindent \textbf{1)} The heterogeneous event attributes have to be represented in form which is adequate as input features for machine learning computations. The categorical attributes (i.e. activity label) are one-hot encoded into a vector $\mathbf{a}^{(i)}$, and the continuous attributes (i.e. timestamp) are min-max scaled to the range of $[0, 1]$ and represented with a scalar $t^{(i)}$. Our framework uses two attributes: the activity label, and the timestamp. We take the time attribute on the scale of seconds granularity and transform it to relative time expressing the duration of events (by the difference of two consecutive event timestamps). This also prevents a particular data leakage from the test set as indicated by \cite{DBLP:journals/corr/abs-2107-01905}. The time attribute value corresponding to any special symbols for the activity label (e.g. \eos{}, \pad{}, \sos{}, and \mask{}) is defined as zero.
\noindent \textbf{2)} For utilising parallel processing resources, the input sequences are batched. This introduces a constraint: sequences in a batch have to be of equal length. Several approaches are possible to ensure that, such as splitting the original sequences into smaller moving-windows of sub-sequences or left/middle/right padding of the original sequences \cite{Lopez-delRio2020}. All approaches may impact the prediction performance of the model: windowing limits the receptive field of the model and padding changes the activity label distribution of the dataset due to the excessive amount of padding needed.

Second, the event attributes have to be fused. Since attributes are perfectly aligned within a trace, we can deal with aligned fusion of the different modalities. \cite{DBLP:journals/dss/HeinrichZJB21,DBLP:conf/enase/SchonigJAJ18,DBLP:conf/caise/TaxVRD17,doi:10.1137/1.9781611976700.59} use feature concatenation, \cite{marlon-lstm-2019} creates  cross-feature embedding of the categorical features offline, \cite{DBLP:conf/emnlp/ZadehCPCM17} applies tensor fusion in the high-dimensional Cartesian space of the feature vectors, \cite{tsai-etal-2019-multimodal} performs crossmodal attention on the concatenated sequence of all modalities, \cite{doi:10.1137/1.9781611975673.14,app11020864} utilise learnable transformations for mapping and weighting the influence of the attributes.

We also apply learnable linear transformations for each feature. The scalar variables (e.g. time attribute) are defined as a rank-zero tensor then broadcasted into a rank-one tensor $E_0\mathbf{t}^{(i)}$\footnote{$E$ stands for the tensor dimension Expansion operation} before the transformations. The resulting vectors are summed to form the distributed embedding vector $\mathbf{z}^{(i)}=\mathbf{a}^{(i)}+E_0\mathbf{t}^{(i)}$ (see Fig.~\ref{figure:framework}).

\subsection{Sequential Deep Learning Models} \label{section:sequential}
\begin{figure}[t]
	\centerline{\includegraphics[width=\columnwidth,keepaspectratio]{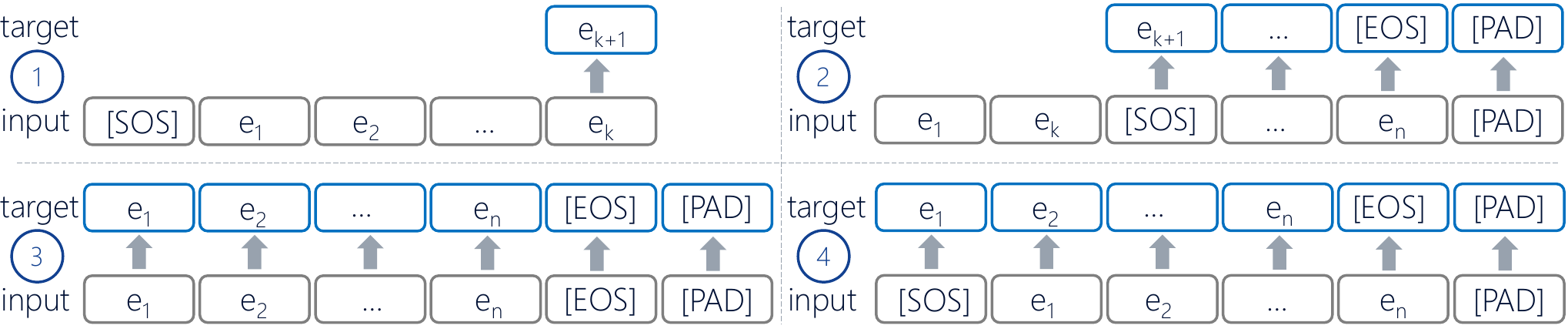}}
	\caption{Input and target variants during training: \circled{1} for LSTM, \circled{2} for AE, AE-GAN and Transformer, \circled{3} for BERT, \circled{4} for GPT and WaveNet.}
	\label{figure:targets}
\end{figure}

\noindent We introduce ML with special emphasis on generative and AutoRegressive (AR) modeling, then we detail several DL architectures that we tested in the framework.

ML seeks to develop methods to automate certain prediction tasks on the basis of observed data. Almost all ML tasks can be formulated as making inferences about missing or latent data from the observed data. To make inferences about unobserved data from the observed data, the learning system needs to make some assumptions; taken together these assumptions constitute a model.
%
%
Learning from data occurs through the transformation of the prior probability distributions (defined before observing the data), into posterior distributions (after observing data) \cite{Ghahramani2015}.

We define an \emph{input space} $\mathcal{X}$ which is a subset of $d$-dimensional real space $\mathbb{R}^d$. We define also a random variable $X$ with probability distribution $P(X)$ which takes values drawn from $\mathcal{X}$. We call the realisations of $X$ feature vectors and noted $x_i$. A generative model describes the marginal distribution over $X$: $P_\Theta(X)$, where samples $x_i$ of $X$ are observed at learning time in a dataset $D$ and the probability distribution depends on some unknown parameter $\Theta$. A generative model family which is important for sequential analysis is the AR one. Here, we fix an ordering of the variables $X_1,X_2, \ldots ,X_n$ and the distribution for the $i$-th random variable depends on the values of all the preceding random variables in the chosen ordering $X_1,X_2, \ldots, X_{i-1}$~\cite{scheduled_sampling}. By the chain rule of probability, we can factorize the joint distribution over the $n$-dimensions as: $P_\Theta(X) = \prod_{i=1}^{n}{P_\Theta(X_i \mid X_1,X_2, \ldots, X_{i-1})}$. AR modeling performs training on dataset $D$ to estimate of the unknown parameter $\Theta$ by maximizing the likelihood under the forward autoregressive factorization: $\max_{\Theta} \log P_\Theta(\mathbf{x}) = \sum_{i=1}^{n} \log P_\Theta(x_i \mid \mathbf{x}_{<i})$. In our scope the \emph{event} $e_i$ from a \emph{trace} $\sigma = \langle e_1, \ldots, e_i, \ldots, e_n \rangle$ (see in Def.~\ref{def:suffix_pred_definition}) is the analogy of $x_i$.

We describe 7 DL architectures (out of which 4 are fist time applied on suffix generation in PM) that all model autoregression but are built up with different building blocks altering training and inference (computation and memory) complexity, model (parameter) size, perspective view (along the sequence), path length (between two positions in the sequence), and parallelisability of operations during training.

\noindent \textbf{LSTM:} \cite{DBLP:journals/dss/EvermannRF17,DBLP:conf/caise/TaxVRD17,marlon-lstm-2019} are pioneer DL solutions applying Recurrent Neural Network (RNN) \cite{RNN} for predictive PM. RNN is an AR model with a feedback loop which allows processing the previous output with the current input, thus making the network stateful, being influenced by the earlier inputs in each step. The simplified equation of a simple recurrent layer is: $\mathbf{z}^{(i)} = \mbox{sigma} (\mathbf{\Theta}_{e}\mathbf{e}^{(i)}+\mathbf{\Theta}_{z}\mathbf{z}^{(i-1)})$\footnote{Where the different $\Theta$s are the learnable parameters}. The hidden state $\mathbf{z}^{(i)}$ depends on the current input $\mathbf{e}^{(i)}$ and its previous hidden state $\mathbf{z}^{(i-1)}$. There is a non-linear dependency (e.g., the sigma function) between them. Long Short Term Memory (LSTM) \cite{LSTM} is a more recent variant which has an additional horizontal residual channel and learnable gating mechanisms for improving on the gradient vanishing during Back Propagation Through Time (BPTT). The first LSTM application to process mining is \cite{DBLP:conf/caise/TaxVRD17} with a single-event target during training. They augment the dataset $D$ with all prefix combinations of traces ($\mathcal{P}$, see in Def.~\ref{def:suffix_pred_definition}) during preprocessing. During training, the the target is the prediction of the sole next event (see \circled{1} on Fig.~\ref{figure:targets}). During inference, there is an open-loop suffix generation of events conditioned on a given prefix, that continues until the \eos{} special symbol or reaching a predefined maximum length: $\hat{\sigma}_{>k}=\verb|rnn|_\Theta(\sigma_{\leq k})$. Equal-length prefixes can be batched together for better utilisation of parallel processing resources.

\noindent \textbf{AE:} An AutoEncoder (AE) is a bottleneck architecture that turns a high-dimensional input into a latent low-dimensional code (encoder), and then performs a reconstruction of the input with this latent code (the decoder) \cite{HinSal06}. The two components are jointly trained together via the reconstruction objective. Sequential autoencoder (i.e. encoder and decoder are both sequential models such as an RNN) is applied in PM for suffix prediction by \cite{doi:10.1137/1.9781611975673.14,doi:10.1137/1.9781611976700.59}. The encoder learns the representation of the prefix, the decoder of the suffix. The encoder maps the prefix into its latent space: $\mathbf{z}=\verb|seq_encoder|_{\Theta_{\mbox{\tiny{enc}}}}(\sigma_{\leq k})$, then the decoder predicts the suffix conditioned on that (i.e. the encoder passes a context vector representing the prefix): $\hat{\sigma}_{>k}=\verb|seq_decoder|_{\Theta_{\mbox{\tiny{dec}}}}(\mathbf{z})$. The dataset $D$ is augmented with all of prefix-suffix combinations $\mathcal{S}$ of traces. The target during the training is to reconstruct the one-step lookahead version of the suffix input (\circled{2} on Fig.~\ref{figure:targets}). The suffix (input and target) can be padded. During inference, after encoding the prefix, there is an open-loop suffix generation of events started with the \sos{} special symbol and until the \eos{} or reaching a predefined maximum length.

\noindent \textbf{AE-GAN:} \cite{doi:10.1137/1.9781611976700.59} extended the sequential AE with a sequence discriminator component which provides feedback about the likeliness of the generated suffixes: $p=\verb|seq_discriminator|_{\Theta_{\mbox{\tiny{disc}}}}(s)$, where $s\sim\mathrm{GumbelSoftmax}(\hat{\sigma}_{>k},\tau)$, and $\tau$ is a temperature parameter which is annealed during training. The architecture setup and the training mechanism follows the Generative Adversarial Network (GAN) \cite{GAN}, although, its objective is not exclusively the minimax game. The discriminator adds an auxiliary adversarial objective to the existing reconstruction one, and it is trained jointly with the autoencoder. In the zero-sum game setup, the objective of the autoencoder is to generate suffixes which follow the distribution of the dataset ($\hat{\sigma}_{>k}\sim\hat{p}_D \leftrightarrow \min \mathbb{E}_{s\sim \mathrm{GS}}[1-\log \verb|seq_discriminator|_{\Theta_{\mbox{\tiny{disc}}}}(s)]$), and the adversarial objective of the discriminator is to discriminate the synthetic ones (i.e. assign $p\approx0 \rightarrow \max \mathbb{E}_{s\sim \mathrm{GS}}[1-\log \verb|seq_discriminator|_{\Theta_{\mbox{\tiny{disc}}}}(s)]$) from the real ones (i.e. assign $p\approx1 \rightarrow \max \mathbb{E}_{\sigma_{>k} \sim \hat{p}_D}[\log \verb|seq_discriminator|_{\Theta_{\mbox{\tiny{disc}}}}(\sigma_{>k})]$).

\noindent The auxiliary adversarial loss for AE is:

\noindent$\mathcal{L}_{\mbox{\tiny{adversarial}}} \allowbreak = \allowbreak 1-\allowbreak \log \verb|seq_discriminator|_{\Theta_{\mbox{\tiny{disc}}}}(s)$. To alleviate the exposure bias by teacher forcing (i.e. during training the target is the sole next event but during inference a suffix is generated), open-loop decoding is applied $90\%$ of the time.

\noindent \textbf{Transformer:} One of the main improvements of the Transformer model is to parallelise the sequence processing. To parallelise the sequence processing (i.e. avoid the recurrence) and improve on the encoder-decoder relation, i.e. access more than a sole context vector to draw global dependencies between input and output, \cite{Transformer} introduces the Transformer. This is a sequential AE architecture with self-attention mechanism in the encoder and decoder, respectively, and cross-attention between the two. Self-attention is an attention mechanism relating different positions of a sequence in order to compute the representation of the sequence. The attention function can be described as mapping a \textit{query} and a set of \textit{key}-\textit{value} pairs to an output, where the \textit{query} $\mathbf{q}$, \textit{keys} $\mathbf{k}$, \textit{values} $\mathbf{v}$, and output $\mathbf{z}$ are all vectors. The output is computed as a weighted sum of the \textit{values}, where the weight assigned to each value is computed by a compatibility/similarity function of the \textit{query} with the corresponding \textit{key}. The sequence of input events are first packed into $\mathbf{Z}^{(0)} = [\mathbf{e}^{(1)}, \cdots, \mathbf{e}^{(n)}]$, and then encoded into contextual representations at different levels of abstraction $\mathbf{Z}^{(l)} = [\mathbf{z}^{(1,l)}, \cdots, \mathbf{z}^{(n,l)}]$ using an $L$-layer Transformer $\mathbf{Z}^{(l)} = \verb|transformer|_{l}(\mathbf{Z}^{(l-1)}), l \in [1,L]$. In each Transformer block, multiple self-attention heads are used to aggregate the output vectors of the previous layer. For the $l$-th Transformer layer, the output of a self-attention head $\mathbf{A}^{(l)}$ is computed via:
\begin{align}
    \mathbf{Q} &= \mathbf{Z}^{(l-1)} \mathbf{\Theta}_Q^{(l)} ,\quad \mathbf{K} = \mathbf{Z}^{(l-1)} \mathbf{\Theta}_K^{(l)},\quad \mathbf{V} = \mathbf{Z}^{(l-1)} \mathbf{\Theta}_V^{(l)} \\
    \mathbf{M}_{ij} &= \begin{cases} 0, &\text{allow to attend} \\ -\infty, &\text{prevent from attending} \end{cases} \label{eq:att:mask} \\
    \mathbf{A}^{(l)} &= \mathrm{softmax}(\frac{\mathbf{Q} \mathbf{K}^{\intercal}}{ \sqrt{d_k}} + \mathbf{M}) \mathbf{V}^{(l)}
\end{align}
where the previous layer's output $\mathbf{Z}^{(l-1)} \in \mathbb{R}^{n \times d_z}$ is linearly projected to a triple of \textit{queries}, \textit{keys} and \textit{values} using parameter matrices $\mathbf{\Theta}_Q^{(l)} , \mathbf{\Theta}_K^{(l)} , \mathbf{\Theta}_V^{(l)} \in \mathbb{R}^{d_z \times d_k}$, respectively, and the mask matrix $\mathbf{M} \in \mathbb{R}^{n \times n}$ determines whether a pair of positions (in the sequence) can be attended to each other to control what context a token can attend to when computing its contextualised representation. The encoder has bidirectional modeling, meaning that the elements of $\mathbf{M}$ are all $0$s, indicating that all the positions have access to each other. The decoder has a left-to-right modeling objective. The representation of each token encodes only the leftward context tokens and itself. This is done by using a triangular matrix for $\mathbf{M}$, where the upper triangular part is set to $-\infty$ and the other elements to $0$. In the encoder-decoder/cross attention layers, the queries come from the previous decoder layer, and the memory keys and values come from the output of the encoder. This allows every position in the decoder to attend over all positions in the input sequence. This mimics the typical encoder-decoder attention mechanisms in sequence-to-sequence models. Self-attention is translation invariant, so its architecture is extended by positional encoding mechanism which could be absolute, or relative. In our framework we apply the absolute positional encoding based on the sinusoidal function. The Transformer has the advantage of simultaneous processing of sequence positions even if its complexity is: $\mathbf{A}^{n \times n} \rightarrow O({n}^2)$. Furthermore, it has a receptive field of $n$ and a maximum path length of $O(1)$. For suffix generations we apply the same prefix-suffix, input-target, and training-inference settings as for the sequential autoencoder $\hat{\sigma}_{>k}=\verb|transformer|_\Theta(\sigma_{\leq k})$. We are the first who apply the Transformer architecture on suffix generation in a multimodal input and multitask prediction setting, \cite{bukhsh2021processtransformer,german-transformer-paper} apply  variants of self-attentional architectures tailored to next event predictions.

\noindent \textbf{GPT:} A Transformer variant named Generative Pre-Training (GPT) has been successfully used for sentence generation in NLP \cite{Radford2018ImprovingLU} and for next-event prediction in PM \cite{app11020864}. It is the decoder block of the Transformer which has a left-to-right modeling capacity. \cite{app11020864} do not model the multimodal event input and does not aim to have multi-target predictions either. In our framework we study GPT's suffix generation performance in the multimodal input and multitarget prediction setting. \circled{4} on Fig.~\ref{figure:targets} visualises the input-target setting during training. There is no need for data augmentation of prefix-suffix splits, this makes the training very effective because any of the sequences/traces can be batched together. During inference, it is also an open-loop suffix generation of events, conditioned on a given prefix and until the \eos{} special symbol or reaching a predefined maximum length $\hat{\sigma}_{>k}=\verb|gpt|_\Theta(\sigma_{\leq k})$.

\noindent \textbf{BERT:} A language representation model called BERT, which stands for Bidirectional Encoder Representations from Transformers is introduced for bidirectional representation by \cite{devlin-etal-2019-bert}. BERT is designed to pre-train deep bidirectional representations from text by jointly conditioning on both left and right context in all layers. To alleviate the unidirectionality constraint it uses a Masked Language Model (MLM) training objective. MLM randomly masks some of the tokens from the input, and the objective is to predict the original masked word based only on its context. Unlike left-to-right language model training, the MLM objective enables the representation to fuse the left and the right context. BERT is based on denoising autoencoding. Specifically, for a sequence $\mathbf{e}$, BERT first constructs a corrupted version $\bar{\mathbf{e}}$ by randomly setting a portion (e.g. $15\%$) of tokens in $\mathbf{e}$ to a special symbol \mask{}. Mathematically, the corruption procedure can be written as $\bar{\mathbf{e}} = (1 - \mathbf{m}) \mathbf{e} + \mathbf{m} \mask{}$, where $\mathbf{m}$ is a binary mask of the same size as $\mathbf{e}$, which uses $1$ to indicate a token will be masked. This is followed by training a Transformer encoder model to reconstruct only the masked tokens in the original text, denoted as $\hat{\mathbf{e}}$, based on the corrupted text $\bar{\mathbf{e}}$. The training objective is to reconstruct $\hat{\mathbf{e}}$ from $\bar{\mathbf{e}}$: $\max_{\theta} \log P_\theta(\hat{\mathbf{e}} \mid \bar{\mathbf{e}}) \approx \sum_{i=1}^{n} m_i \log P_\theta(e_i \mid \bar{\mathbf{e}})$, where $m_i = 1$ indicates $e_i$ is masked. As emphasized by the $\approx$ sign, BERT factorizes the joint conditional probability $P(\hat{\mathbf{e}} \mid \bar{\mathbf{e}})$ based on an independence assumption that all masked tokens $\hat{\mathbf{e}}$ are separately reconstructed. \circled{3} on Fig.~\ref{figure:targets} visualises the input-target setup during training. There is no need for data augmentation of prefix-suffix splits, this makes the training very effective because any of the sequences/traces can be batched together. To enhance BERT for sequence generation, we applied a probabilistic mask portion (PMLM) \cite{PMLM} instead of a fixed percentage. In each training step, the masking ratio is sampled from the uniform distribution $U(1,n)$. Hence, during the course of training, all masked permutations of the sequence is seen. During inference, the initial condition is that all the suffix positions are masked, then during the loop, all suffix positions are visited one-by-one in a random order. The token at the predicted position is added then to the context for the next step until all suffix positions are visited. During evaluation the sub-sequence from left, up to the first \eos{} is resulted $\hat{\sigma}_{>k}=\verb|bert|_\Theta(\sigma_{\leq k})$.

\noindent \textbf{WaveNet:} \cite{oord2016wavenet} is a generative model which has been introduced for signal time series. It is a fully ($1$-D) Convolutional Neural Network (CNN), where the stacked convolutional layers are causal and dilated. There are no pooling layers in the network, and the output of the model has the same time dimensionality as the input: \circled{4} on Fig.~\ref{figure:targets} visualises the input-target setting during training which is the same setup as for GPT. At training time, the conditional predictions for all positions can be made in parallel because all positions of ground truth are known; the inference procedure is sequential as of all the previous AR models $\hat{\sigma}_{>k}=\verb|wavenet|_\Theta(\sigma_{\leq k})$. One of the problems of causal
convolutions is that they require many layers, or large filters to increase the receptive field. The authors apply dilated convolutions to increase the receptive field by orders of magnitude, without greatly increasing computational cost. In each layer the dilations are increasing by a factor of two, hence, the receptive field is: $r=2^{L-1}k$, where $L$ is the number of layers and $k$ is the convolutional filter size. We are the first who apply a CNN architecture on suffix generation in a multimodal input and multitask prediction setting, \cite{10.1007/978-3-030-66498-5_24} applies a variant of it tailored to process outcome prediction and \cite{8786066} is tailored to next event prediction.

\subsection{Generator} \label{section:generator}
The generator component offers multitask predictions. For each event in the sequence, categorical and numerical variables are predicted, respectively. For the categorical \emph{activity label} variable, the readout provides a vector of logits $\mathbf{\hat{a}^{(i)}}$. The logits are transformed into likelihoods by the softmax function ($\mathrm{softmax}(\mathbf{\hat{a}}^{(i)})_c=\frac{\exp(\mathbf{\hat{a}}_c^{(i)})}{\sum_{c}\exp(\mathbf{\hat{a}}_c^{(i)})}$). There are several methods to decode the discrete token out of this categorical distribution of $\mathbf{\hat{a}^{(i)}}$. The most commonly used method is to select the most likely category \cite{DBLP:journals/dss/HeinrichZJB21,DBLP:conf/enase/SchonigJAJ18,DBLP:conf/caise/TaxVRD17}, that is also called as argmax or greedy search. Beam search, an algorithmic breadth-first search technique, can yield better quality of sequences with the cost of increased memory and computation \cite{doi:10.1137/1.9781611976700.59}. Different forms of stochastic sampling (from the categorical distribution) can be used also \cite{Radford2018ImprovingLU,DBLP:journals/dss/EvermannRF17,marlon-lstm-2019}. We use the most common greedy search solution for the experiments.

The proposed framework provides multi-objective optimisation. Learning for the categorical features is via the categorical cross-entropy loss (i.e. of the ground truth $\mathbf{a^{(i)}}$ and the predicted $\mathbf{\hat{a}^{(i)}}$): $\mathcal{L}_{\mbox{\tiny{cross-entropy}}}=-\mathbf{a^{(i)}} \cdot \log \mathbf{\hat{a}^{(i)}}$. The loss function is the average of such errors over all items in the sequence.
Learning for the continuous features is via the squared error (i.e. of the ground truth $t^{(i)}$  and the predicted $\hat{t}^{(i)}$): $\mathcal{L}_{\mbox{\tiny{squared error}}}=(\hat{t}^{(i)} - t^{(i)})^2$. The loss function is the average of such errors over all items in the sequence.

The final loss is a weighted sum of the individual losses. During inference the sequence generation is in a loop; a step is always conditioned on the previously generated context up to the \eos{} or predefined maximum length. We set that maximum to the length of the longest trace in the dataset which is a pragmatic choice.

\section{Experimental Results}
\label{sec:results}
\subsection{Evaluation Measures}
To evaluate the performance of suffix prediction (in the viewpoint of the sequence of categorical activity labels), we use Damerau-Levenstein distance (DL). This metric measures the quality of the predicted suffix of a \emph{trace} by adding swapping operation to the set of operations used by regular Levenstein distance. Given two activity sequences $s_1$ and $s_2$. For example, $s_1 = \langle \mathbf{a^{(>k)}}, \ldots ,\mathbf{a^{(n)}}\rangle$, and $s_2 = \langle \mathbf{\hat{a}^{(>k)}}, \ldots ,\mathbf{\hat{a}^{(m)}}\rangle$. We consider the following similarity: $\mathrm{DLS}(s_1,s_2) = 1-\frac{\mathrm{DL}(s_1,s_2)}{\max(\mathrm{len}(s_1),\mathrm{len}(s_2))}$, where $\mathrm{len}(s)$ is the length of $s$. $\mathrm{DLS} \in [0,1]$, and it is $1.0$ when two sequences are the same and $0.0$ when two sequences contain completely different elements. Also, we compute the absolute error between the ground truth remaining time and the predicted remaining time for each predicted and ground truth suffixes. Next, we average these numbers for evaluation instances, and report Mean Absolute Error (MAE).

\subsection{Datasets}
\begin{figure}[t!]
	\centerline{\includegraphics[width=\columnwidth,keepaspectratio]{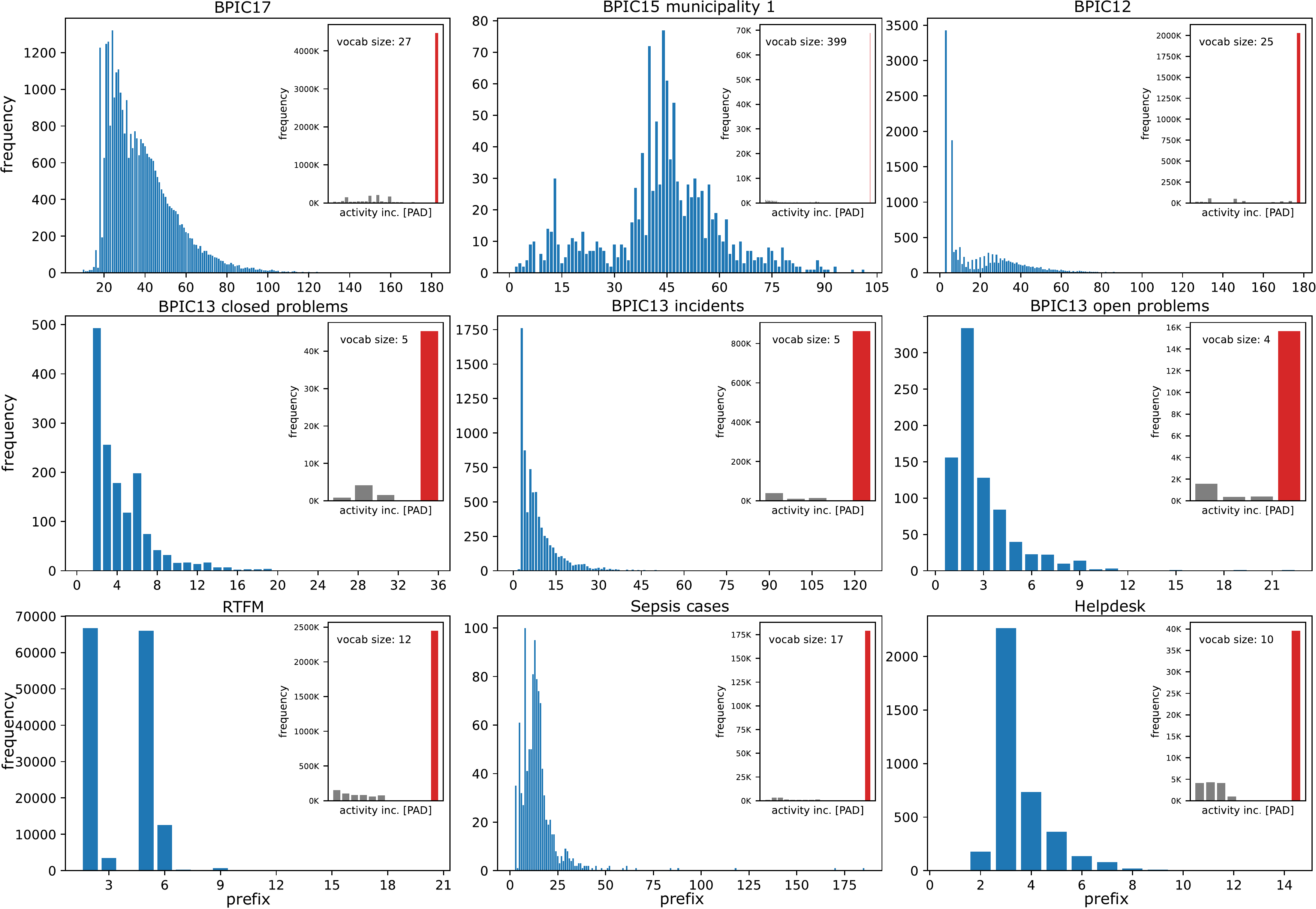}}
	\caption{Case length and activity (including [PAD] symbol) distributions (per dataset).}
	\label{figure:distributions}
\end{figure}
We evaluate the performance of the models on
$9$ real-life datasets (which are very commonly used in the PM community)\footnote{Publicly available at \url{https://data.4tu.nl/search?q=:keyword:\%20\%22Task\%20Force\%20on\%20Process\%20Mining\%22}.} without prepossessing. Fig.~\ref{figure:distributions} shows the case length and activity distributions for all the datasets. The case length distribution of all datasets is heavily skewed. This brings a challenging situation for sequential prediction tasks. The models have to represent the underrepresented long traces, in the long tail of the distributions. The embedded activity distribution plots include the added \pad{} special symbol (red bar) and show the extreme relative frequency of \pad{}.
We apply splitting into training and evaluation sets by 8:2 ratio after shuffling of traces, and use the $80\%$ of the traces for training, and the remaining $20\%$ traces for evaluation. We use the exact same subsets for all experiments.

\subsection{Experimental Setup}
The proposed framework is implemented\footnote{\url{http://github.com/smartjourneymining/sequential-deep-learning-models}} in Python $3.7$, PyTorch $1.9$, and CUDA $10.2$ on a Linux server using an NVIDIA V100 $16$-GB GPU.
Throughout all experiments we keep an equal number of $4$ layers with $128$ latent vector size of the sequential component, plus an embedding and generator with the same latent size. We train models for $400$ epochs and apply early stopping if we observe no more improvement on the evaluation set for 50 iterations. We use Adam as an optimization algorithm for the proposed framework with learning rate $1\mathrm{e}{-4}$. To improve on generalisation we apply a dropout rate of $0.3$.

\subsection{Results}
\begin{table}[tb]
    \caption{Average DLS results (\textbf{best} and \textit{worst} performers per dataset).}\label{tab:dls}
    \scriptsize
    \centerline{\input{table_dls_33}}
\end{table}
As Table~\ref{tab:dls} shows all the models except BERT perform on a comparable level. The reason might be that BERT is more sensitive to the skewness because of its non-AR training objective. Furthermore, masked language modeling (PMLM) might require more training epochs (since it approximates all perturbations) for suffix prediction.
AR models perform well on datasets with shorter traces and smaller vocabulary such as the Helpdesk and RTFM logs. Those logs are also relatively more much structured. BPI17 and Sepsis cases appear to be the most challenging datasets. Together with BPI12, those also have the longest traces.

\begin{figure*}[t!]
	\centerline{\includegraphics[width=\textwidth,keepaspectratio]{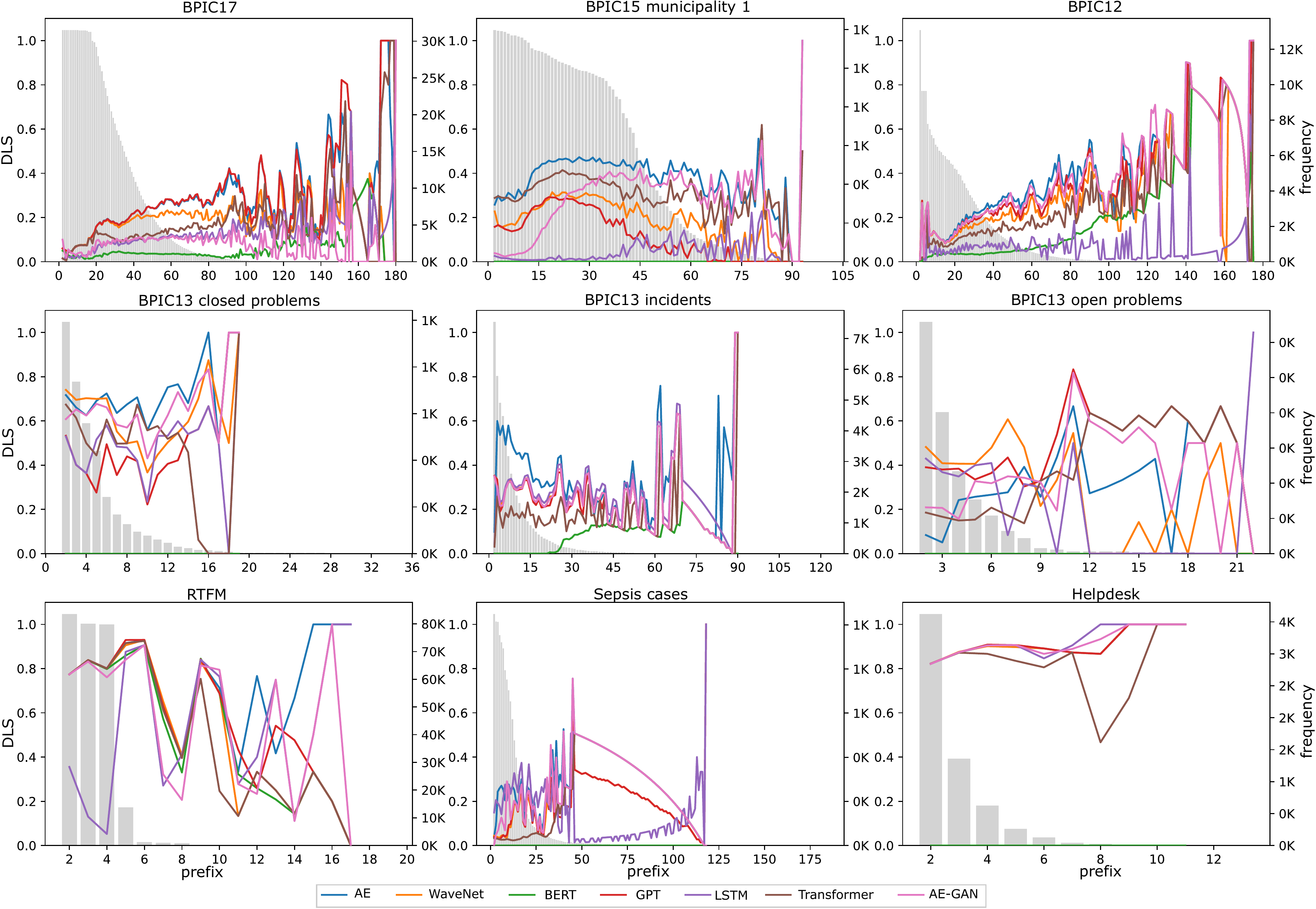}}
	\caption{DLS and frequency values over all prefix combinations per dataset.}
	\label{figure:dls}
\end{figure*}

Fig.~\ref{figure:dls} visualises the DLS of the suffix generations results for different prefix lengths (x-axis) starting from length two. We visualise the performance change of the model with a line chart (left y-axis) and the frequency of prefixes of a certain length in the data with a bar chart (right y-axis).
The intuition would be that by increasing prefix size the DLS generally increases but it is empirically different. The bar chart aims to shed the light of the underlying phenomena. It could be connected to the number of traces in the dataset which are equal or longer than the given prefix length. Due to the non uniform trace length distribution, the count of traces is monotonically decreasing given the increasing prefix length. The slope of that decrease is small for shorter prefixes: in that interval the DLS generally increases. That is generally followed by a large drop in the count possibly due to the skewness: that results in performance degradation of the models in general. The models tend to be biased towards predicting shorter suffixes because there are much less longer traces in the dataset. The DLS calculation heavily punishes that bias. However, there are perfect DLS scores for the prefix sizes which are very close to the longest trace(s). That may be the result of overfitting given the infinitesimal variance among those extremely few traces.


It is interesting that there are results missing for some prefix lengths even though these prefixes do exist in the data (bar chart). This is due to the training-evaluation random split: in the evaluation subset there are no corresponding traces with the same length.
Thus, a sole, average DLS metric seems to be not always informative enough. For example, for the RTFM dataset GPT has the highest average DLS but for the prefix length of 12, it is the worst performing.

\begin{table}[tb]
    \caption{Average MAE results (\textbf{best} and \textit{worst} performers per dataset).}\label{tab:mae}
    \scriptsize
    \centerline{\input{table_mae_33}}
\end{table}
Table~\ref{tab:mae} shows that all the models perform on a comparable level regarding average MAE of the time prediction. LSTM, BERT, and WaveNet perform worse which might be because of the the limited receptive field in case of WaveNet and LSTM. BERT might needs more training iterations because of the randomness introduced in its MLM.
\begin{figure*}[t!]
	\centerline{\includegraphics[width=\textwidth,keepaspectratio]{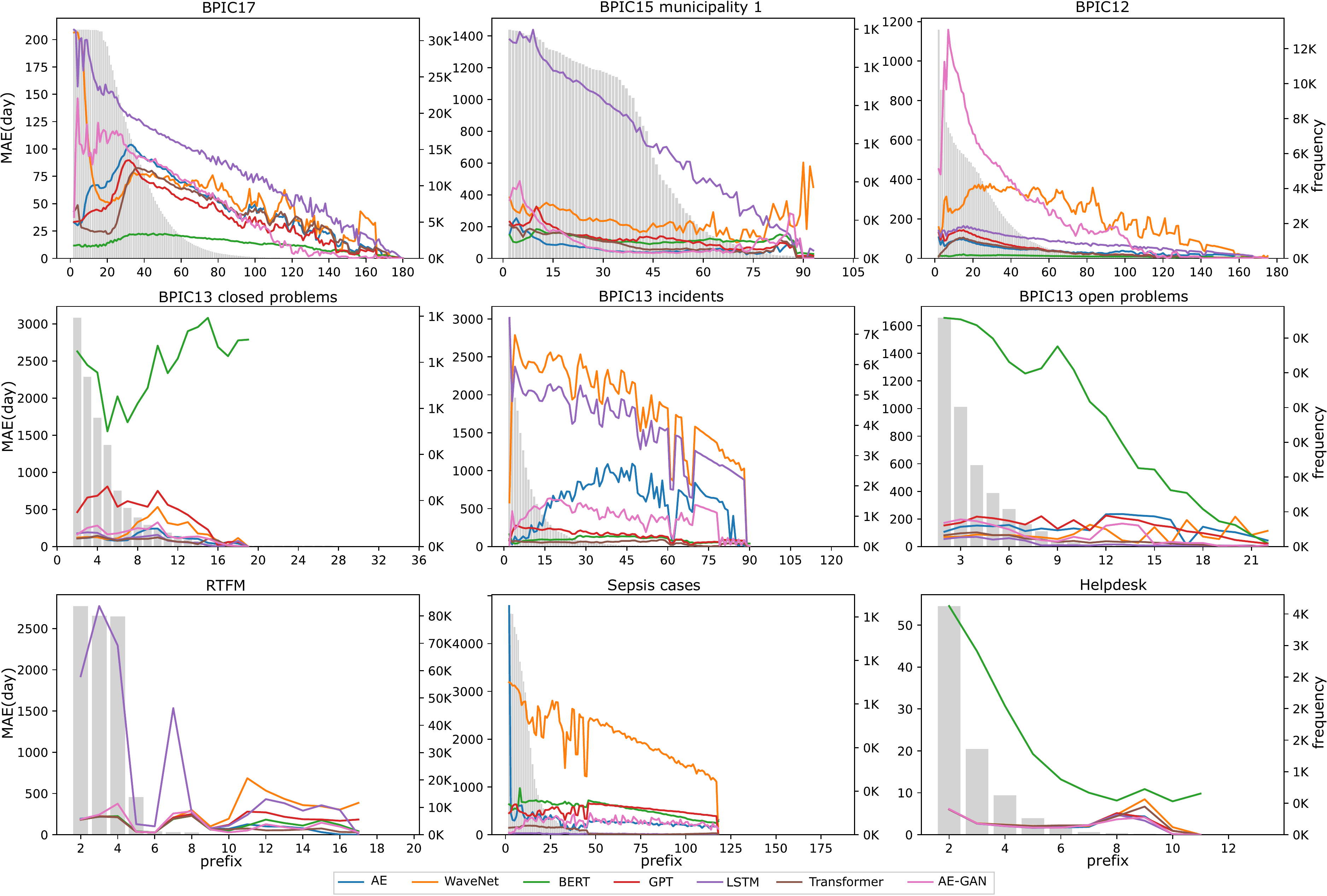}}
	\caption{MAE (in days) and frequency values over all prefix combinations per dataset.}
	\label{figure:mae}
\end{figure*}
Fig.~\ref{figure:mae} visualises the MAE of suffix generation results with two charts in an overlay for all prefix combinations (x-axis) starting from length two. The line chart (left y-axis) visualises the the performance change of the models over the changing prefix size. The bar chart (right y-axis)  plots the number of traces in the dataset which are equal or longer than the given prefix length. Due to the non uniform trance length distribution, the count of traces is monotonically decreasing given the increasing prefix length. The intuition is that by increasing prefix size the MAE generally decreases since there are less remaining time predictions to be made. This seems to be empirically supported.

\section{Conclusions}

We investigated how 7 sequential DL architectures (out of which 4 have never been applied on suffix generation in PM) perform for predicting the suffix of multi-modal sequences. Evaluating the architectures across all prefix lengths showed that none of them is one-size-fits-all and that with increasing lengths, the prediction performance for some prefixes fluctuates heavily with substantial drops in performance. These results have implications on how performance of DL models should be reported. Solely reporting the average DLS for suffix prediction seems to be inadequate for comparing DL architectures and for assessing their suitability in practice. Our hypothesis is that this effect is due to the skewness of the length distribution for traces in event log, which prevents the direct application of successful DL architectures for NLP tasks (in particular, BERT) in our experimental setting. In future work, also the impact of other event log properties, e.g., as proposed in~\cite{DBLP:journals/dss/HeinrichZJB21,DBLP:journals/bise/KratschMRS21}, should be investigated for suffix prediction. Furthermore, controlled experiments (e.g., on simulated data from process models \cite{DBLP:conf/bpm/Burattin16}) could be carried out to identify the impact of specific data properties. Finally, our experiments should be extended with hyperparameter tuning for each architecture to avoid the influence of manual parameter choices.

\textbf{Acknowledgement.} This work is part of the Smart Journey Mining project, which is funded by the Research Council of Norway (project no. 312198).

\bibliographystyle{ACM-Reference-Format}
\bibliography{literature}

\end{document}

%% file: table_dls_33.tex
\begin{tabularx}{\columnwidth}{@{\hspace{0\tabcolsep}}l@{\hspace{1\tabcolsep}}l@{\hspace{1\tabcolsep}}l@{\hspace{1\tabcolsep}}l@{\hspace{1\tabcolsep}}l@{\hspace{1\tabcolsep}}l@{\hspace{1\tabcolsep}}l@{\hspace{1\tabcolsep}}l@{\hspace{0\tabcolsep}}}
\toprule
model/dataset &      AE$\uparrow$ & WaveNet$\uparrow$ &    BERT$\uparrow$ &     GPT$\uparrow$ &     LSTM$\uparrow$ & Transformer$\uparrow$ &  AE-GAN$\uparrow$ \\
\midrule
BPI17               &  \textbf{0.1417} &  0.1326 &  \textit{0.0306} &  0.1412 &  0.0593 &           0.0931 &  0.0697 \\
BPI15 municipality 1                             &  \textbf{0.4035} &  0.2393 &  \textit{0.0000} &  0.1919 &  0.0276 &           0.3356 &  0.2375 \\
BPI12                   &  \textbf{0.1737} &  0.1518 &  \textit{0.0333} &  0.1647 &  0.0677 &           0.1004 &  0.1666 \\
BPI13 closed problems   &  \textbf{0.6820} &  0.6729 &  \textit{0.0000} &  0.4236 &  0.4721 &           0.5762 &  0.6271 \\
BPI13 incidents         &  \textbf{0.4363} &  0.2894 &  \textit{0.0057} &  0.2859 &  0.2610 &           0.1631 &  0.2863 \\
BPI13 open problems     &  0.1562 &  \textbf{0.4372} &  \textit{0.0000} &  0.3879 &  0.3703 &           0.1887 &  0.2378 \\
RTFM &  0.8236 &  0.8218 &  0.8107 &  \textbf{0.8263} &  \textit{0.3708} &           0.8230 &  0.7975 \\
Sepsis cases    &  0.2169 &  0.0892 &  \textit{0.0000} &  0.0924 &  \textbf{0.2218} &           0.0391 &  0.1396 \\
Helpdesk                             &  0.8593 &  0.8599 &  \textit{0.0000} &  \textbf{0.8609} &  0.8599 &           0.8471 &  0.8599 \\
\bottomrule
\end{tabularx}

%% file: table_mae_33.tex
\begin{tabularx}{\columnwidth}{@{\hspace{0\tabcolsep}}l@{\hspace{1\tabcolsep}}l@{\hspace{1\tabcolsep}}l@{\hspace{1\tabcolsep}}l@{\hspace{1\tabcolsep}}l@{\hspace{1\tabcolsep}}l@{\hspace{1\tabcolsep}}l@{\hspace{1\tabcolsep}}l@{\hspace{0\tabcolsep}}}
\toprule
model/dataset &      AE$\downarrow$ & WaveNet$\downarrow$ &    BERT$\downarrow$ &     GPT$\downarrow$ &     LSTM$\downarrow$ & Transformer$\downarrow$ &  AE-GAN$\downarrow$ \\
\midrule
BPI17               &   69.51 &    90.54 &    \textbf{15.60} &   54.53 &   \textit{152.08} &            43.72 &  100.19 \\
BPI15 municipality 1                             &   \textbf{85.24} &   259.54 &   126.83 &  162.04 &  \textit{1038.16} &           122.67 &  143.94 \\
BPI12                   &   73.59 &   286.61 &    \textbf{14.57} &  104.42 &   133.45 &            66.11 &  \textit{650.32} \\
BPI13 closed problems   &  117.42 &   154.92 &  \textit{2282.33} &  596.65 &   153.50 &           \textbf{110.53} &  208.08 \\
BPI13 incidents         &  215.51 &  \textit{2193.19} &    82.99 &  213.65 &  2173.84 &            \textbf{67.64} &  370.01 \\
BPI13 open problems     &  129.79 &    76.00 &  \textit{1543.89} &  173.31 &    \textbf{55.15} &            83.00 &  161.10 \\
RTFM &  167.66 &   164.59 &   166.51 &  164.05 &  \textit{1748.88} &           \textbf{161.93} &  2015 \\
Sepsis cases     &  735.04 &  \textit{2771.08} &   654.68 &  496.06 &    \textbf{28.45} &           161.69 &  187.12 \\
Helpdesk                             &    \textbf{3.83} &     3.95 &    \textit{43.57} &    3.88 &     3.84 &             3.93 &    3.88 \\
\bottomrule
\end{tabularx}

%% file: sample-sigconf.bbl

\begin{thebibliography}{30}


\ifx \showCODEN    \undefined \def \showCODEN     #1{\unskip}     \fi
\ifx \showDOI      \undefined \def \showDOI       #1{#1}\fi
\ifx \showISBNx    \undefined \def \showISBNx     #1{\unskip}     \fi
\ifx \showISBNxiii \undefined \def \showISBNxiii  #1{\unskip}     \fi
\ifx \showISSN     \undefined \def \showISSN      #1{\unskip}     \fi
\ifx \showLCCN     \undefined \def \showLCCN      #1{\unskip}     \fi
\ifx \shownote     \undefined \def \shownote      #1{#1}          \fi
\ifx \showarticletitle \undefined \def \showarticletitle #1{#1}   \fi
\ifx \showURL      \undefined \def \showURL       {\relax}        \fi
\providecommand\bibfield[2]{#2}
\providecommand\bibinfo[2]{#2}
\providecommand\natexlab[1]{#1}
\providecommand\showeprint[2][]{arXiv:#2}

\bibitem[\protect\citeauthoryear{Bengio, Vinyals, Jaitly, and Shazeer}{Bengio
  et~al\mbox{.}}{2015}]%
        {scheduled_sampling}
\bibfield{author}{\bibinfo{person}{Samy Bengio}, \bibinfo{person}{Oriol
  Vinyals}, \bibinfo{person}{Navdeep Jaitly}, {and} \bibinfo{person}{Noam
  Shazeer}.} \bibinfo{year}{2015}\natexlab{}.
\newblock \showarticletitle{Scheduled Sampling for Sequence Prediction with
  Recurrent Neural Networks}.
\newblock In \bibinfo{booktitle}{\emph{Advances in Neural Information
  Processing Systems 28}}, \bibfield{editor}{\bibinfo{person}{C.~Cortes},
  \bibinfo{person}{N.~D. Lawrence}, \bibinfo{person}{D.~D. Lee},
  \bibinfo{person}{M.~Sugiyama}, {and} \bibinfo{person}{R.~Garnett}} (Eds.).
  \bibinfo{publisher}{Curran Associates, Inc.}, \bibinfo{pages}{1171--1179}.
\newblock


\bibitem[\protect\citeauthoryear{Bukhsh, Saeed, and Dijkman}{Bukhsh
  et~al\mbox{.}}{2021}]%
        {bukhsh2021processtransformer}
\bibfield{author}{\bibinfo{person}{Zaharah~A. Bukhsh}, \bibinfo{person}{Aaqib
  Saeed}, {and} \bibinfo{person}{Remco~M. Dijkman}.}
  \bibinfo{year}{2021}\natexlab{}.
\newblock \bibinfo{title}{ProcessTransformer: Predictive Business Process
  Monitoring with Transformer Network}.
\newblock
\newblock
\showeprint[arxiv]{cs.LG/2104.00721}


\bibitem[\protect\citeauthoryear{Burattin}{Burattin}{2016}]%
        {DBLP:conf/bpm/Burattin16}
\bibfield{author}{\bibinfo{person}{Andrea Burattin}.}
  \bibinfo{year}{2016}\natexlab{}.
\newblock \showarticletitle{{PLG2:} Multiperspective Process Randomization with
  Online and Offline Simulations}. In \bibinfo{booktitle}{\emph{Proceedings of
  the {BPM} Demo Track 2016 Co-located with the 14th International Conference
  on Business Process Management {(BPM} 2016), Rio de Janeiro, Brazil,
  September 21, 2016}} \emph{(\bibinfo{series}{{CEUR} Workshop Proceedings})},
  \bibfield{editor}{\bibinfo{person}{Leonardo Azevedo} {and}
  \bibinfo{person}{Cristina Cabanillas}} (Eds.), Vol.~\bibinfo{volume}{1789}.
  \bibinfo{publisher}{CEUR-WS.org}, \bibinfo{pages}{1--6}.
\newblock
\urldef\tempurl%
\url{http://ceur-ws.org/Vol-1789/bpm-demo-2016-paper1.pdf}
\showURL{%
\tempurl}


\bibitem[\protect\citeauthoryear{Camargo, Dumas, and
  Gonz{\'a}lez-Rojas}{Camargo et~al\mbox{.}}{2019}]%
        {marlon-lstm-2019}
\bibfield{author}{\bibinfo{person}{Manuel Camargo}, \bibinfo{person}{Marlon
  Dumas}, {and} \bibinfo{person}{Oscar Gonz{\'a}lez-Rojas}.}
  \bibinfo{year}{2019}\natexlab{}.
\newblock \showarticletitle{Learning Accurate LSTM Models of Business
  Processes}. In \bibinfo{booktitle}{\emph{Business Process Management}},
  \bibfield{editor}{\bibinfo{person}{Thomas Hildebrandt},
  \bibinfo{person}{Boudewijn~F. van Dongen}, \bibinfo{person}{Maximilian
  R{\"o}glinger}, {and} \bibinfo{person}{Jan Mendling}} (Eds.).
  \bibinfo{publisher}{Springer International Publishing},
  \bibinfo{address}{Cham}, \bibinfo{pages}{286--302}.
\newblock
\showISBNx{978-3-030-26619-6}


\bibitem[\protect\citeauthoryear{del Rio et~al\mbox{.}}{del Rio
  et~al\mbox{.}}{2020}]%
        {Lopez-delRio2020}
\bibfield{author}{\bibinfo{person}{Lopez del Rio} {et~al\mbox{.}}}
  \bibinfo{year}{2020}\natexlab{}.
\newblock \showarticletitle{Effect of sequence padding on the performance of
  deep learning models in archaeal protein functional prediction}.
\newblock \bibinfo{journal}{\emph{Sci Rep}} \bibinfo{volume}{10},
  \bibinfo{number}{1} (\bibinfo{year}{2020}), \bibinfo{pages}{14634}.
\newblock
\showISSN{2045-2322}


\bibitem[\protect\citeauthoryear{Devlin, Chang, Lee, and Toutanova}{Devlin
  et~al\mbox{.}}{2019}]%
        {devlin-etal-2019-bert}
\bibfield{author}{\bibinfo{person}{Jacob Devlin}, \bibinfo{person}{Ming-Wei
  Chang}, \bibinfo{person}{Kenton Lee}, {and} \bibinfo{person}{Kristina
  Toutanova}.} \bibinfo{year}{2019}\natexlab{}.
\newblock \showarticletitle{{BERT}: Pre-training of Deep Bidirectional
  Transformers for Language Understanding}. In
  \bibinfo{booktitle}{\emph{NAACL}}. \bibinfo{pages}{4171--4186}.
\newblock


\bibitem[\protect\citeauthoryear{Evermann, Rehse, and Fettke}{Evermann
  et~al\mbox{.}}{2017}]%
        {DBLP:journals/dss/EvermannRF17}
\bibfield{author}{\bibinfo{person}{Joerg Evermann},
  \bibinfo{person}{Jana{-}Rebecca Rehse}, {and} \bibinfo{person}{Peter
  Fettke}.} \bibinfo{year}{2017}\natexlab{}.
\newblock \showarticletitle{Predicting process behaviour using deep learning}.
\newblock \bibinfo{journal}{\emph{Decis. Support Syst.}}  \bibinfo{volume}{100}
  (\bibinfo{year}{2017}), \bibinfo{pages}{129--140}.
\newblock


\bibitem[\protect\citeauthoryear{Ghahramani}{Ghahramani}{2015}]%
        {Ghahramani2015}
\bibfield{author}{\bibinfo{person}{Zoubin Ghahramani}.}
  \bibinfo{year}{2015}\natexlab{}.
\newblock \showarticletitle{Probabilistic machine learning and artificial
  intelligence}.
\newblock \bibinfo{journal}{\emph{Nature}} \bibinfo{volume}{521},
  \bibinfo{number}{7553} (\bibinfo{year}{2015}), \bibinfo{pages}{452--459}.
\newblock
\showISSN{1476-4687}
\urldef\tempurl%
\url{https://doi.org/10.1038/nature14541}
\showDOI{\tempurl}


\bibitem[\protect\citeauthoryear{Goodfellow, Pouget-Abadie, Mirza, Xu,
  Warde-Farley, Ozair, Courville, and Bengio}{Goodfellow et~al\mbox{.}}{2014}]%
        {GAN}
\bibfield{author}{\bibinfo{person}{Ian Goodfellow}, \bibinfo{person}{Jean
  Pouget-Abadie}, \bibinfo{person}{Mehdi Mirza}, \bibinfo{person}{Bing Xu},
  \bibinfo{person}{David Warde-Farley}, \bibinfo{person}{Sherjil Ozair},
  \bibinfo{person}{Aaron Courville}, {and} \bibinfo{person}{Yoshua Bengio}.}
  \bibinfo{year}{2014}\natexlab{}.
\newblock \showarticletitle{{G}enerative {A}dversarial {N}ets}.
\newblock In \bibinfo{booktitle}{\emph{Advances in Neural Information
  Processing Systems 27}}.
\newblock


\bibitem[\protect\citeauthoryear{Heinrich, Zschech, Janiesch, and
  Bonin}{Heinrich et~al\mbox{.}}{2021}]%
        {DBLP:journals/dss/HeinrichZJB21}
\bibfield{author}{\bibinfo{person}{Kai Heinrich}, \bibinfo{person}{Patrick
  Zschech}, \bibinfo{person}{Christian Janiesch}, {and} \bibinfo{person}{Markus
  Bonin}.} \bibinfo{year}{2021}\natexlab{}.
\newblock \showarticletitle{Process data properties matter: Introducing gated
  convolutional neural networks {(GCNN)} and key-value-predict attention
  networks {(KVP)} for next event prediction with deep learning}.
\newblock \bibinfo{journal}{\emph{Decis Support Syst}}  \bibinfo{volume}{143}
  (\bibinfo{year}{2021}), \bibinfo{pages}{113494}.
\newblock


\bibitem[\protect\citeauthoryear{Hinton and Salakhutdinov}{Hinton and
  Salakhutdinov}{2006}]%
        {HinSal06}
\bibfield{author}{\bibinfo{person}{Geoffrey Hinton} {and}
  \bibinfo{person}{Ruslan Salakhutdinov}.} \bibinfo{year}{2006}\natexlab{}.
\newblock \showarticletitle{Reducing the Dimensionality of Data with Neural
  Networks}.
\newblock \bibinfo{journal}{\emph{Science}} \bibinfo{volume}{313},
  \bibinfo{number}{5786} (\bibinfo{year}{2006}), \bibinfo{pages}{504 -- 507}.
\newblock


\bibitem[\protect\citeauthoryear{Hochreiter and Schmidhuber}{Hochreiter and
  Schmidhuber}{1997}]%
        {LSTM}
\bibfield{author}{\bibinfo{person}{Sepp Hochreiter} {and}
  \bibinfo{person}{J\"{u}rgen Schmidhuber}.} \bibinfo{year}{1997}\natexlab{}.
\newblock \showarticletitle{{Long} {Short}-{Term} {Memory}}.
\newblock \bibinfo{journal}{\emph{Neural Comput.}} \bibinfo{volume}{9},
  \bibinfo{number}{8} (\bibinfo{date}{Nov.} \bibinfo{year}{1997}),
  \bibinfo{pages}{1735--1780}.
\newblock
\showISSN{0899-7667}
\urldef\tempurl%
\url{https://doi.org/10.1162/neco.1997.9.8.1735}
\showDOI{\tempurl}


\bibitem[\protect\citeauthoryear{Jordan}{Jordan}{1986}]%
        {RNN}
\bibfield{author}{\bibinfo{person}{Michael~I. Jordan}.}
  \bibinfo{year}{1986}\natexlab{}.
\newblock \showarticletitle{Attractor Dynamics and Parallelism in a
  Connectionist Sequential Machine}. In \bibinfo{booktitle}{\emph{{P}roceedings
  of the Eighth Annual Conference of the Cognitive Science Society}}.
  \bibinfo{publisher}{Hillsdale, NJ: Erlbaum}, \bibinfo{pages}{531--546}.
\newblock


\bibitem[\protect\citeauthoryear{Kratsch et~al\mbox{.}}{Kratsch
  et~al\mbox{.}}{2021}]%
        {DBLP:journals/bise/KratschMRS21}
\bibfield{author}{\bibinfo{person}{Wolfgang Kratsch} {et~al\mbox{.}}}
  \bibinfo{year}{2021}\natexlab{}.
\newblock \showarticletitle{Machine Learning in Business Process Monitoring:
  {A} Comparison of Deep Learning and Classical Approaches Used for Outcome
  Prediction}.
\newblock \bibinfo{journal}{\emph{BISE}}  \bibinfo{volume}{63}
  (\bibinfo{year}{2021}), \bibinfo{pages}{261--276}.
\newblock


\bibitem[\protect\citeauthoryear{Liao, Jiang, and Liu}{Liao
  et~al\mbox{.}}{2020}]%
        {PMLM}
\bibfield{author}{\bibinfo{person}{Yi Liao}, \bibinfo{person}{Xin Jiang}, {and}
  \bibinfo{person}{Qun Liu}.} \bibinfo{year}{2020}\natexlab{}.
\newblock \showarticletitle{Probabilistically {Masked} {Language} {Model}
  Capable of Autoregressive Generation in Arbitrary Word Order}. In
  \bibinfo{booktitle}{\emph{ACL proceedings}}. \bibinfo{pages}{263--274}.
\newblock


\bibitem[\protect\citeauthoryear{Lin, Wen, and Wang}{Lin et~al\mbox{.}}{2019}]%
        {doi:10.1137/1.9781611975673.14}
\bibfield{author}{\bibinfo{person}{Li Lin}, \bibinfo{person}{Lijie Wen}, {and}
  \bibinfo{person}{Jianmin Wang}.} \bibinfo{year}{2019}\natexlab{}.
\newblock \showarticletitle{{MM-Pred}: A Deep Predictive Model for
  Multi-attribute Event Sequence}. In \bibinfo{booktitle}{\emph{SDM19}}.
  \bibinfo{pages}{118--126}.
\newblock


\bibitem[\protect\citeauthoryear{Moon, Park, and Jeong}{Moon
  et~al\mbox{.}}{2021}]%
        {app11020864}
\bibfield{author}{\bibinfo{person}{Junhyung Moon}, \bibinfo{person}{Gyuyoung
  Park}, {and} \bibinfo{person}{Jongpil Jeong}.}
  \bibinfo{year}{2021}\natexlab{}.
\newblock \showarticletitle{{POP-ON}: Prediction of Process Using One-Way
  Language Model Based on {NLP} Approach}.
\newblock \bibinfo{journal}{\emph{Applied Sciences}} \bibinfo{volume}{11},
  \bibinfo{number}{2} (\bibinfo{year}{2021}).
\newblock
\showISSN{2076-3417}


\bibitem[\protect\citeauthoryear{Neu, Lahann, and Fettke}{Neu
  et~al\mbox{.}}{2021}]%
        {Neu2021}
\bibfield{author}{\bibinfo{person}{Dominic~A. Neu}, \bibinfo{person}{Johannes
  Lahann}, {and} \bibinfo{person}{Peter Fettke}.}
  \bibinfo{year}{2021}\natexlab{}.
\newblock \showarticletitle{A systematic literature review on state-of-the-art
  deep learning methods for process prediction}.
\newblock \bibinfo{journal}{\emph{Artificial Intelligence Review}}
  (\bibinfo{year}{2021}).
\newblock


\bibitem[\protect\citeauthoryear{Pasquadibisceglie, Appice, Castellano, and
  Malerba}{Pasquadibisceglie et~al\mbox{.}}{2019}]%
        {8786066}
\bibfield{author}{\bibinfo{person}{Vincenzo Pasquadibisceglie},
  \bibinfo{person}{Annalisa Appice}, \bibinfo{person}{Giovanna Castellano},
  {and} \bibinfo{person}{Donato Malerba}.} \bibinfo{year}{2019}\natexlab{}.
\newblock \showarticletitle{Using Convolutional Neural Networks for Predictive
  Process Analytics}. In \bibinfo{booktitle}{\emph{2019 International
  Conference on Process Mining (ICPM)}}. \bibinfo{pages}{129--136}.
\newblock
\urldef\tempurl%
\url{https://doi.org/10.1109/ICPM.2019.00028}
\showDOI{\tempurl}


\bibitem[\protect\citeauthoryear{Philipp, Jacob, Robert, and Beyerer}{Philipp
  et~al\mbox{.}}{2020}]%
        {german-transformer-paper}
\bibfield{author}{\bibinfo{person}{Patrick Philipp}, \bibinfo{person}{Ruben
  Jacob}, \bibinfo{person}{Sebastian Robert}, {and} \bibinfo{person}{Jürgen
  Beyerer}.} \bibinfo{year}{2020}\natexlab{}.
\newblock \showarticletitle{Predictive Analysis of Business Processes Using
  Neural Networks with Attention Mechanism}. In \bibinfo{booktitle}{\emph{2020
  International Conference on Artificial Intelligence in Information and
  Communication (ICAIIC)}}. \bibinfo{pages}{225--230}.
\newblock
\urldef\tempurl%
\url{https://doi.org/10.1109/ICAIIC48513.2020.9065057}
\showDOI{\tempurl}


\bibitem[\protect\citeauthoryear{Radford and Narasimhan}{Radford and
  Narasimhan}{2018}]%
        {Radford2018ImprovingLU}
\bibfield{author}{\bibinfo{person}{Alec Radford} {and} \bibinfo{person}{Karthik
  Narasimhan}.} \bibinfo{year}{2018}\natexlab{}.
\newblock \showarticletitle{Improving Language Understanding by Generative
  Pre-Training}.
\newblock


\bibitem[\protect\citeauthoryear{Sch{\"{o}}nig, Jasinski, Ackermann,
  et~al\mbox{.}}{Sch{\"{o}}nig et~al\mbox{.}}{2018}]%
        {DBLP:conf/enase/SchonigJAJ18}
\bibfield{author}{\bibinfo{person}{Stefan Sch{\"{o}}nig},
  \bibinfo{person}{Richard Jasinski}, \bibinfo{person}{Lars Ackermann},
  {et~al\mbox{.}}} \bibinfo{year}{2018}\natexlab{}.
\newblock \showarticletitle{Deep Learning Process Prediction with Discrete and
  Continuous Data Features}. In \bibinfo{booktitle}{\emph{{ENASE 2018}}}.
  \bibinfo{pages}{314–319}.
\newblock


\bibitem[\protect\citeauthoryear{Tax, Verenich, Rosa, and Dumas}{Tax
  et~al\mbox{.}}{2017}]%
        {DBLP:conf/caise/TaxVRD17}
\bibfield{author}{\bibinfo{person}{Niek Tax}, \bibinfo{person}{Ilya Verenich},
  \bibinfo{person}{Marcello~La Rosa}, {and} \bibinfo{person}{Marlon Dumas}.}
  \bibinfo{year}{2017}\natexlab{}.
\newblock \showarticletitle{Predictive Business Process Monitoring with {LSTM}
  Neural Networks}. In \bibinfo{booktitle}{\emph{CAiSE 2017}}
  \emph{(\bibinfo{series}{LNCS})}, Vol.~\bibinfo{volume}{10253}.
  \bibinfo{pages}{477--492}.
\newblock


\bibitem[\protect\citeauthoryear{Taymouri, Rosa, and Erfani}{Taymouri
  et~al\mbox{.}}{2021}]%
        {doi:10.1137/1.9781611976700.59}
\bibfield{author}{\bibinfo{person}{Farbod Taymouri},
  \bibinfo{person}{Marcello~La Rosa}, {and} \bibinfo{person}{Sarah~M. Erfani}.}
  \bibinfo{year}{2021}\natexlab{}.
\newblock \showarticletitle{A Deep Adversarial Model for Suffix and Remaining
  Time Prediction of Event Sequences}. In \bibinfo{booktitle}{\emph{SDM21}}.
  \bibinfo{pages}{522--530}.
\newblock


\bibitem[\protect\citeauthoryear{Tsai, Bai, Liang, et~al\mbox{.}}{Tsai
  et~al\mbox{.}}{2019}]%
        {tsai-etal-2019-multimodal}
\bibfield{author}{\bibinfo{person}{Yao-Hung~Hubert Tsai},
  \bibinfo{person}{Shaojie Bai}, \bibinfo{person}{Paul~Pu Liang},
  {et~al\mbox{.}}} \bibinfo{year}{2019}\natexlab{}.
\newblock \showarticletitle{Multimodal {Transformer} for Unaligned Multimodal
  Language Sequences}. In \bibinfo{booktitle}{\emph{ACL proceedings}}.
  \bibinfo{pages}{6558--6569}.
\newblock


\bibitem[\protect\citeauthoryear{van~den Oord et~al\mbox{.}}{van~den Oord
  et~al\mbox{.}}{2016}]%
        {oord2016wavenet}
\bibfield{author}{\bibinfo{person}{Aaron van~den Oord} {et~al\mbox{.}}}
  \bibinfo{year}{2016}\natexlab{}.
\newblock \showarticletitle{{WaveNet}: A Generative Model for Raw Audio}. In
  \bibinfo{booktitle}{\emph{Arxiv}}.
\newblock
\urldef\tempurl%
\url{https://arxiv.org/abs/1609.03499}
\showURL{%
\tempurl}


\bibitem[\protect\citeauthoryear{Vaswani, Shazeer, Parmar,
  et~al\mbox{.}}{Vaswani et~al\mbox{.}}{2017}]%
        {Transformer}
\bibfield{author}{\bibinfo{person}{Ashish Vaswani}, \bibinfo{person}{Noam
  Shazeer}, \bibinfo{person}{Niki Parmar}, {et~al\mbox{.}}}
  \bibinfo{year}{2017}\natexlab{}.
\newblock \showarticletitle{Attention is All you Need}. In
  \bibinfo{booktitle}{\emph{Advances in Neural Information Processing
  Systems}}, Vol.~\bibinfo{volume}{30}.
\newblock


\bibitem[\protect\citeauthoryear{Weytjens and De~Weerdt}{Weytjens and
  De~Weerdt}{2020}]%
        {10.1007/978-3-030-66498-5_24}
\bibfield{author}{\bibinfo{person}{Hans Weytjens} {and} \bibinfo{person}{Jochen
  De~Weerdt}.} \bibinfo{year}{2020}\natexlab{}.
\newblock \showarticletitle{Process Outcome Prediction: CNN vs. LSTM (with
  Attention)}. In \bibinfo{booktitle}{\emph{Business Process Management
  Workshops}}, \bibfield{editor}{\bibinfo{person}{Adela Del R{\'i}o~Ortega},
  \bibinfo{person}{Henrik Leopold}, {and} \bibinfo{person}{Fl{\'a}via~Maria
  Santoro}} (Eds.). \bibinfo{publisher}{Springer International Publishing},
  \bibinfo{address}{Cham}, \bibinfo{pages}{321--333}.
\newblock
\showISBNx{978-3-030-66498-5}


\bibitem[\protect\citeauthoryear{Weytjens and {De Weerdt}}{Weytjens and {De
  Weerdt}}{2021}]%
        {DBLP:journals/corr/abs-2107-01905}
\bibfield{author}{\bibinfo{person}{Hans Weytjens} {and} \bibinfo{person}{Jochen
  {De Weerdt}}.} \bibinfo{year}{2021}\natexlab{}.
\newblock \showarticletitle{Creating Unbiased Public Benchmark Datasets with
  Data Leakage Prevention for Predictive Process Monitoring}. In
  \bibinfo{booktitle}{\emph{Arxiv}}.
\newblock
\urldef\tempurl%
\url{https://arxiv.org/abs/2107.01905}
\showURL{%
\tempurl}


\bibitem[\protect\citeauthoryear{Zadeh, Chen, Poria, Cambria, and
  Morency}{Zadeh et~al\mbox{.}}{2017}]%
        {DBLP:conf/emnlp/ZadehCPCM17}
\bibfield{author}{\bibinfo{person}{Amir Zadeh}, \bibinfo{person}{Minghai Chen},
  \bibinfo{person}{Soujanya Poria}, \bibinfo{person}{Erik Cambria}, {and}
  \bibinfo{person}{Louis-Philippe Morency}.} \bibinfo{year}{2017}\natexlab{}.
\newblock \showarticletitle{{Tensor} {Fusion} {Network} for Multimodal
  Sentiment Analysis}. In \bibinfo{booktitle}{\emph{{EMNLP}}}.
  \bibinfo{pages}{1103--1114}.
\newblock


\end{thebibliography}
